\documentclass[10pt, a4paper]{article}
\usepackage{lrec2022} % this is the new LREC2022 Style
\usepackage{multibib}
\newcites{languageresource}{Language Resources}
\usepackage{graphicx}
\usepackage{tabularx}
\usepackage{soul}
\usepackage{subcaption}
\usepackage{expex}
\usepackage{titlesec}
\titleformat{\section}{\normalfont\large\bfseries\center}{\thesection.}{1em}{}
\titleformat{\subsection}{\normalfont\SmallTitleFont\bfseries\raggedright}{\thesubsection.}{1em}{}
\titleformat{\subsubsection}{\normalfont\normalsize\bfseries\raggedright}{\thesubsubsection.}{1em}{}
\renewcommand\thesection{\arabic{section}}
\renewcommand\thesubsection{\thesection.\arabic{subsection}}
\renewcommand\thesubsubsection{\thesubsection.\arabic{subsubsection}}
\usepackage{epstopdf}
\usepackage[utf8]{inputenc}

\usepackage{hyperref}
\usepackage{xstring}

\usepackage{color}

\title{Lutma: a Frame-Making Tool for Collaborative FrameNet Development}

\name{Tiago Timponi Torrent\textsuperscript{1,2}, Arthur Lorenzi\textsuperscript{1}, Ely Edison da Silva Matos\textsuperscript{1}, \\
\bf \large{Frederico Belcavello\textsuperscript{1}, Marcelo Viridiano\textsuperscript{1}, Maucha Andrade Gamonal\textsuperscript{3}}}

\address{\textsuperscript{1} FrameNet Brasil Lab, Graduate Program in Linguistics, Federal University of Juiz de Fora \\
\textsuperscript{2} Brazilian National Council for Scientific and Technological Development – CNPq \\ 
\textsuperscript{3} LETra, Graduate Program in Linguistic Studies, Federal University of Minas Gerais \\
         \{tiago.torrent, ely.matos, fred.belcavello\}@ufjf.br, \{arthur.lorenzi, barros.marcelo \}@estudante.ufjf.br,  \\ mgamonal@ufmg.br}
         
\abstract{
This paper presents Lutma, a collaborative, semi-constrained, tutorial-based tool for contributing frames and lexical units to the Global FrameNet initiative. The tool parameterizes the process of frame creation, avoiding consistency violations and promoting the integration of frames contributed by the community with existing frames. Lutma is structured in a wizard-like fashion so as to provide users with text and video tutorials relevant for each step in the frame creation process. We argue that this tool will allow for a sensible expansion of FrameNet coverage in terms of both languages and cultural perspectives encoded by them, positioning frames as a viable alternative for representing perspective in language models.
 \\ \newline  }

\begin{document}

\maketitleabstract

\section{Introduction}

As we have claimed elsewhere, ``language can be vague, messy and variable because humans cooperate while using it'' \cite{torrent-2021}, and communicative cooperation critically involves sharing culture, and perspectives on it and inviting our interlocutors to reconstrue them as needed. We claim that computational models of language should embrace the characteristics of human language and not try to overcome them by the mathematical manipulation of linguistic form patterns extracted from large datasets alone \cite{bender-koller-2020-climbing}. Moreover, we claim that the FrameNet model \cite{fillmore2010frames,torrent2018towards} is a good candidate for representing those characteristics, provided that it is extended to cover more languages and dialects.

Doing so is not trivial if one considers how FrameNets have been built so far – see, among others, \newcite{fillmore2003framenet},\newcite{ohara2004japanese}, \newcite{torrent2013behind} and \newcite{dannells2021swedish}. All of them relied on a relatively small team of intensively trained linguists, who invested a considerable amount of time building machine-readable frames and associating linguistic material with them based on corpus evidence representing a small number of languages and an even smaller number of variants within one same language. Due to this \emph{modus operandi}, FrameNets usually present limited coverage, which is frequently pointed out by NLP practitioners as a reason for not using them in their applications. Nonetheless, the knowledge accumulated in the past three decades of FrameNet development allows for a methodological turn, presented in this paper in the form of a software tool: Lutma \footnote{\url{https://lutma.frame.net.br}}. 

Lutma is semi-constrained, tutorial-based tool for fostering a community of distributed frame builders who will be able to enrich FrameNet with more diverse perspectives grounded on their own languages and language variants, enhancing its coverage and representativeness. In the remainder of the paper we start, in section \ref{sec:case}, by making the case in favor of a larger, multilingual and multidialectal FrameNet, by contrasting FrameNet with Large Language Models (LLMs). Next, we present Lutma in section \ref{sec:lutma}. Section \ref{sec:example} presents an example of culturally grounded frame creation. Finally, section \ref{sec:limitations} closes the paper by presenting the current limitations and future developments planed for collaborative FrameNet building.

%more representative Such an enriched FrameNet could then be presented as an alternative to enrich language models with cognitively and culturally relevant features so that they improve their performance while at the same time minimizing their biases and potential harms. In a nutshell, the main idea we have been pursuing is that a collaborative multilingual and multimodal FrameNet, built by a diverse community, can reframe language models so that they no longer remove language from its cognitive, cultural, social and multimodal context.

%In this paper, we defend that this idea not only expands on existing discussions about perspective, but we also present our first step into this endeavor: the development of Lutma, a community-based frame creation tool.

\section{The Case for a Collaborative Frame Building Tool}
\label{sec:case}

In recent years, discussions about perspectives have flourished in NLP, motivated by a variety of reasons, among which the problems of reducing multiple labels into a single ground-truth label, the ambiguous nature of many NLP tasks and bias encoded on large language models \cite{aroyo2015truth,artstein2008inter,bender2021dangers}. Integrating multiple perspectives into models, however, is far from trivial, as specific model characteristics must be taken into consideration.

In white-box systems, one can take many different approaches to prevent bias and guarantee representation of multiple perspectives, namely through pre- and post-processing, but also by altering the models' internals \cite{ntoutsi2020bias}. For that reason, one can argue that they are particularly useful in cases where curators and developers want to make sure that the model -- or the dataset -- are not encoding bias \cite{criado2019digital}.

In the case of black-box models, the strategies used in supervised and unsupervised models to introduce multiple perspectives are very distinct. For the former – represented by Machine Translation (MT), Question Answering (QA), Named Entity Recognition (NER) and many other models – \newcite{basile2021toward} defend the adoption of \textit{data perspectivism}, moving away from gold standard datasets and instead adopting different points of view for each object in the data. In practical terms, this variation is obtained by assigning the annotation task of a single data record to multiple human subjects. The way those different perspectives are integrated into a model determines whether the implementation follows a \textit{weak} or a \textit{strong perspectivist} approach. Any system that aggregates – using majority, average, etc. – those different perspectives into a single label is considered to follow the \textit{weak perspectivist} method. When, instead, the model is adapted to handle the outputs from multiple annotators, it is classified as a \textit{strong prespectivist} one. This added complexity, and consequently the amount of work, is outweighed by the ethical principles being followed, and in many cases, also leads to better performance \cite{basile2021toward}. 

Unsupervised methods, on the other hand, pose different types of challenges. We turn to them next.

\subsection{Perspective in LLMs}
\label{sec:llms}

Unsupervised methods are primarily represented by LLMs, such as BERT \cite{devlin2018bert}, GPT-3 \cite{raffel2019exploring} and T5 \cite{brown2020language}, which cannot integrate multiple perspectives into their training by expanding the number of gold standard labels per object. Rather, pre-existing bias in the data needs to be addressed during corpus curation and preprocessing.

\newcite{bender2021dangers} state that models trained on large and uncurated text datasets encode biases that lead to unethical technology, also calling for investment to curate those datasets as a means to avoid such biases. \newcite{rogers2021changing} claims that curation already takes place in the datasets feeding language models, and poses the question of what type of curation would be most effective to avoid harmful biases and improve models' abilities to understand language.

%Finally, in regards to the data used for training, documentation is an important action to be taken, with the goal of facilitating future analyses that could reveal biases \cite{bender2018data}. That, of course, does not actively impact training, but improves the discussions about pre-existing biases in the community.

One key aspect of LLMs is that, regardless of whether data curation takes place or not, perspective understanding is treated as a byproduct of the mathematical manipulation of linguistic form. In other words, the machine's role is to ``figure out'' different perspectives solely from variations in form, while the researcher is responsible for making sure that the training data is representative of those perspectives. Within this framework, whether or not multiple points of view are being considered by the model depends on the dataset size and the quality of the data sample, much like the overall model performance.

We, in turn, claim that, instead of manipulating only input data, NLP systems should integrate a cognitively and culturally-oriented model that represents alternative perspectives on the meaning of linguistic forms. We also claim FrameNet to be the most suitable model for doing that.

\subsection{Perspective in FrameNet}
\label{sec:framenet}

Perspective is a core aspect of Frame Semantics \cite{Fillmore1982,fillmore-85}. Directly related to the classic Fillmorean proposition that \textit{meanings are relativized to scenes} is the idea that different perspectives may be taken on that scene \cite{fillmore1977case}. The classic example revolves around the perspectives on the commercial transaction event. \newcite{fillmore1977case} demonstrates that different English verbs – namely \textit{buy, sell, cost} and \textit{charge} – adopt the perspectives of the different participants in the scene – Buyer, Seller, Goods and Money. Each perspective structures the scene in a particular fashion, so that some of the participants that are core to one perspective – e.g. Buyer and Goods in the perspective lexicalized in \textit{buy.v} – may not be central to others – e.g. in the perspective lexicalized in \textit{charge.v}, to which the Seller and the Money are central.

As the implementation of Frame Semantics, FrameNet models frames in terms of the elements in them, the coreness status of those element and the relations established between frames. The model also includes the lexical items evoking the frames. In this context, there are three aspects of FrameNet that may contribute to enrich and diversify language models:

\textbf{Cognitively-based:} FrameNet was initially proposed as lexical database inspired in Frame Semantics \cite{ruppenhofer2016framenet}. Lexical units are linked to frames, which in turn are linked to other frames. According to Frame Semantics, in order to understand a single frame, one needs to understand the structure in which it fits \cite{petruck_frame_1996}. Frames are schematic representations of concepts based on recurring experiences against which the meanings of lexical units are relativized \cite{fillmore1977case}. This means that, instead of representing words with a single vector, or a list of senses, they can be associated with multiple activation patterns in the network. It also means, as demonstrated by \newcite{10.3389/fpsyg.2022.838441}, that FrameNet structure captures contextual information, namely commonsense knowledge.

\textbf{Socially contextualized:} Since frames are schematic representations of concepts, they are also used to represent socially construed entities and events, for example. The existence of specific types of frame relations also facilitates the creation of frames that may represent particular views on the same event. For example, research on Japanese noun-modification constructions by \newcite{matsumoto2010interactional} demonstrates that the societal grounding of the interaction and the purpose of the discourse influence on the grammar of noun-modification, claiming that interactional frames play a central role in the comprehension of those constructions.

\textbf{Multilingual:} Research on Frame Semantics adopting a contrastive multilingual approach has demonstrated that different languages may lexicalize different perspectives on a given scene, one of the parade examples being the study of verbs of emotion in Spanish vs. English \cite{Subirats-Petruck2003}. Because there are framenets under development for a number of languages, those differences are also captured via either the Global FrameNet Shared Annotation task \cite{torrent2018multilingual} or the Multilingual FrameNet database alignment \cite{GILARDI18.11,baker-lorenzi-2020-exploring}. Such efforts allow for the construction of a single database supporting lexical units from any language, but at the same time not restricted to universally applicable frames. Even if some culturally specific frames are created, they can still be linked to the global network of frames. One of the challenges in this undertaking is to merge  resources that were built independently for years. Nonetheless, for those working with low-resource languages (LRLs), Global FrameNet and Multilingual FrameNet are of great help, since they allow users to focus on frame-evoking units in their language, instead of modeling frames and their relations. This possibility of focusing on the language itself rather than on the underlying frame structure can be an important tool to reduce the gap of LRLs in NLP \cite{magueresse2020low,cruz2020establishing,lakew2020low}. 

% \textbf{Multimodal:} Because human communication is inherently multimodal \cite{kress2020reading} - that is, it relies not only on verbal language, but also on gesture and other semiosis, such as videos and pictures - FrameNet has been expanded beyond lexical units. First, by exploring the continuity between those same lexical units and constructions (as in traditional Construction Grammar theories). Then, by modeling images and subtitles from videos as actual frame-evoking entities. A language model trained with data of various modes would better mimic the way humans handle language and could be an advantage even for downstream tasks that only deal with textual data.

Despite the three features described above, FrameNet still lacks coverage in terms of both number of languages and cognitive domains included in the model. Therefore, a fourth feature must be pursued if FrameNet is to be presented as an alternative to representing perspective in language models:

\textbf{Community-based:} Expanding contributions to FrameNet language resources is the main solution to increase the speed at which those resources evolve. And although some challenges come with a community-based approach, it brings an overall positive balance to research pursuits. Having a global community increases the possibility that contributors from varied backgrounds work together. This directly impacts the quality of the resource, as more languages will be supported and also new frames from specific cultural backgrounds will be created.

Implementing this feature is the purpose of Lutma, which is presented next.

\section{Lutma: a Frame-Maker Tool}
\label{sec:lutma}

Lutma is one of the steps towards building an extended FrameNet resource in which language is not isolated from human cognition and social backgrounds. The project is part of the Global FrameNet effort, a collaboration between labs and affiliate researchers of twelve different countries, with the goal of facilitating the sharing of findings and research data, as well as building partnerships for the development of novel research.

The way Lutma differentiates itself from the various tools used by the FrameNets around the world is that it has two design goals not shared by the others: first, that frames and LUs must have a clear indication of the languages/cultures in which they belong; second, that the user experience must be aimed towards people interested in FrameNet, but with little or no training on frame creation. These goals are both aligned with the idea of building a cognitively-based and collaborative language resource.

The main challenge of extending FrameNet lies on the fact that, until recent, it has been mostly reliant on specialists. To address this challenge, frame creation in Lutma follows a linear approach, much like a wizard, where to advance, users are asked a certain amount of information about the records. The idea of having separate steps for different pieces of data during the process also allows the system to run consistency and redundancy checks, making sure that users who are not experienced with the concepts can create frames with adequate quality.

\begin{figure*}[ht]
  \includegraphics[width=\linewidth]{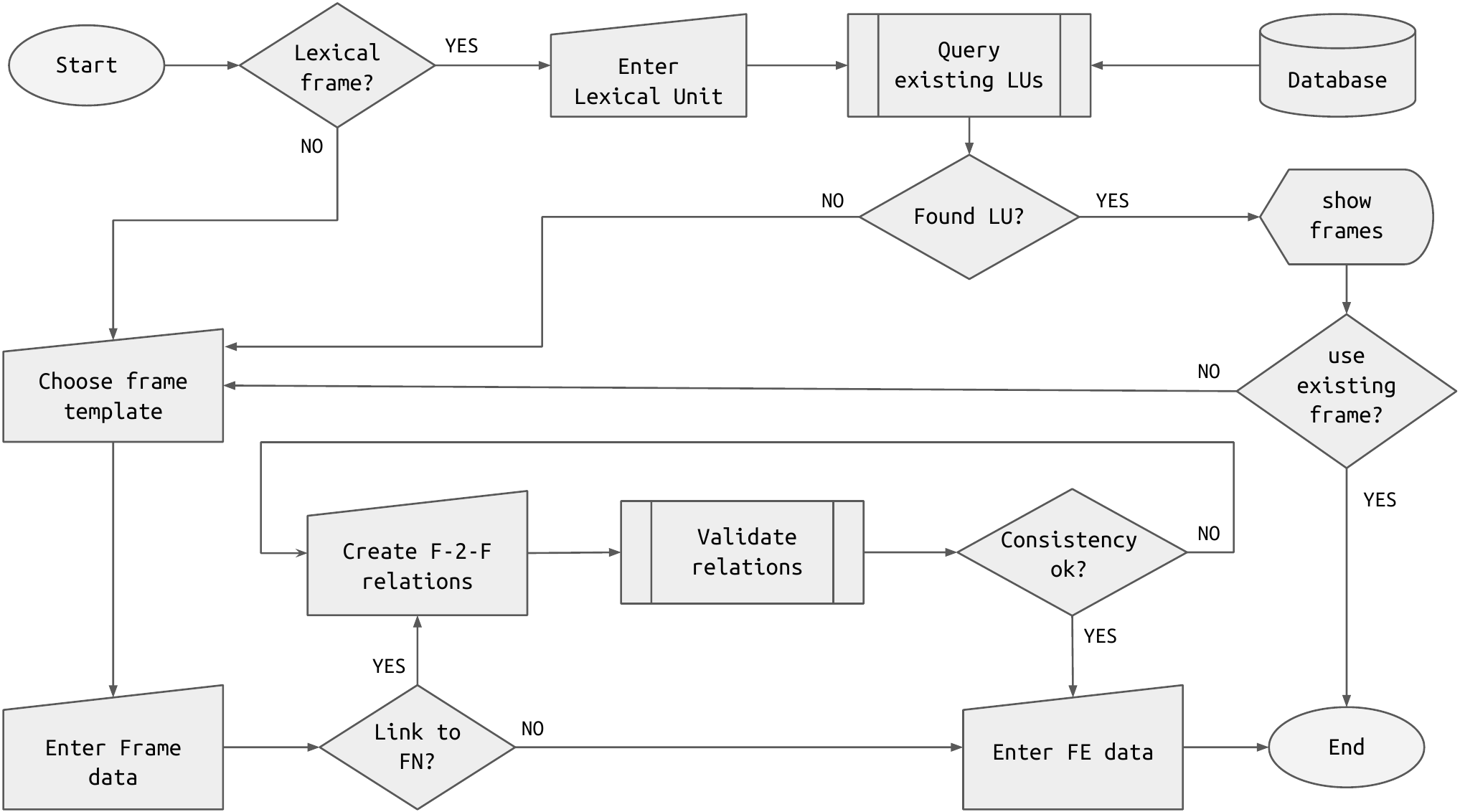}
  \caption{Lutma's frame creation flow. The diagram encompasses the creation of lexical and non-lexical frames.}
  \label{fig:flowchart}
\end{figure*}

Frame creation is separated into two execution flows, one for lexical and the other for non-lexical frames. Both are presented in the flow diagram in Figure \ref{fig:flowchart}. The first one starts with a lemma search: the user inputs the system with the part-of-speech and language of a lemma that will evoke the frame they want to create. The system checks if this lemma already exists in the database and, if not, searches for synonyms using Open Multilingual WordNet (OMWN) \cite{bond2013linking}. If any of the synonyms is an LU in the database, the evoked frames are displayed and the user can decide whether a new frame needs to be created or if a new LU will be created for an existing frame. A second check is executed, but instead of looking for synonyms, Lutma searches for words in other languages with similar spelling in the same OMWN synset. This is a last measure to prevent redundancy, taking advantage of the existing data in other languages, although the multilingual synsets are limited to a small subset of languages.

When a new frame needs to be created, the user is guided to the next screen where they select its root type. They can choose from event, entity, relation, attribute, state or undefined (when it doesn't fit any of the previous). The selected type is used by the system to make suggestions of possible frame elements during the process, e. g., ``Direction'' or ``Material'' for relation and entity frames, respectively. In the next step, the user needs to fill out frame names and definitions and once again the system checks for duplicate names and if the name follows certain standards, such as scenario frame names ending with ``\_scenario'' or state frames having names following the ``Being\_x'' or ``x\_state'' patterns. 

In the following screen, users can create relations between their new frame and existing frames in Lutma's database. When creating those relations, the system allows users to choose whether they want to map FEs from the other frame into the frame (in the case of an inheritance relation, this is required). After storing those relations, the next step consists of creating FEs. At this point, the user's frame may already have some FEs because of the frame relations, but Lutma still suggests more FEs based on the frame type. The rest of the FEs can be manually created and at least one is required to proceed. In the final step, before a summary is shown, the system asks for FE relation information. The creation summary displays all of the information related to the frame and to finish creation users must provide an example sentence for the lemma that was informed at the beginning of the process, as well as inform if it incorporates one of the FEs. After that, the frame is registered in the database.

The flow for non-lexical frames shares most of the steps with the lexical one. The first difference is the absence of a lemma search (because the frame will not be evoked by lexical material). The second is the need to inform the frame language in frame type selection step. We opted for this solution to guarantee that those frames would not be treated as universal, despite the fact that they can be associated to multiple languages.

These two execution flows separated into steps, along with the automatic quality checks run by the system, improve user experience by reducing the chances that a mistake will be made. However, when trying to build a community around this type of resource, we expect users with varying knowledge about FrameNets. For that reason, Lutma also integrates tutorials into its interface. Those tutorials can be categorized into two types, those about the system's interface, that explain how users can achieve certain goals, and those about Frame Semantics and FrameNet. The former are automatically displayed at the first time an user logs in to Lutma, while the latter can be accessed using the interface elements at any step.

These theoretical and practical tutorials found in all system screens can be further divided into two types. The simpler ones are displayed in the form of dialog boxes that are rendered every time a user clicks on one of the UI elements. For example, when searching for a lexical unit in the database, a user can click on the "Enter lemma" label of the search field to open a dialog card explaining what is the definition of lemma used in Lutma. Those small texts are useful for users that want to remember how certain concepts are defined in FrameNet. They can also be useful to users with different backgrounds in Linguistics, facilitating the comprehension of how those same concepts are represented in Lutma.

When users need more information than provided in the dialog boxes, the system presents a link to a video tutorial on the specific subject. These videos are always paired with one of the previously mentioned dialog boxes as a way of presenting a longer discussion for the concept. They are also tailored to a broader audience and explain essential concepts in a simple manner, using various examples. In total, six videos were produced, ranging from three to nine minutes in duration. Most videos address more than one topic relevant to the process of frame creation and, because of that, are linked to different parts of the systems, but effectively covering all of the dialog boxes. For future work, these tutorials could be expanded even further, including more topics and references to relevant publications.

Last but not least, to make sure that the data created by collaborators will benefit other contributors or projects, we opted for a copyleft license, namely GPLv3 \footnote{\url{www.gnu.org/licenses/gpl-3.0.html}}. With this licensing scheme, not only the data is made accessible for any interested party, but improvements made outside of its ecosystem can be potentially reintegrated into the database.

Since the project is quite new, there are still limitations that need to be addressed before a full release for the community. The final section discusses those and summarizes our contributions. Before turning to them, though, we present an example demonstrating how culturally specific frames can be created in Lutma and integrated to the existing FrameNet frames.

\section{The Brazilian Way Frame}
\label{sec:example}

To provide an example of how Lutma may aid in the expansion of FrameNet to include culturally grounded frames, while linking them to the existing database, we present the creation of the \texttt{Brazilian\_way} frame, evoked by LUs such as \emph{jeitinho.n}, \emph{malandragem.n} and \emph{gingado.n} in Brazilian Portuguese (br-pt). Those LUs can be literally translated into English (en) as \emph{`little way'}, \emph{`trickery'} and \emph{`waddle'}, respectively. Their culturally grounded meaning is quite different though.

According to \newcite{damatta_1986}, \emph{jeitinho} is characterized as the space Brazilians find between what one can do and what one cannot do within a normative system. Such a system can be institutionalized in the Judiciary, or may correspond to implicit social norms that should be followed by everyone. When the concept of \emph{jeitinho} is brought into play, one seeks to solve some private problem by adopting some behavior that makes the solution easier and/or faster. Such a behavior is inadequate under the strict observation of the norm regulating the problem-solving task. Nonetheless, by bringing \emph{jeitinho} into play, such inadequacy is relativized.  

The fact that the main lexical unit in this frame is in the diminutive form is not coincidental. The br-pt expression \emph{dar um jeito} corresponds roughly to the en verb \emph{fix} in sentences like \emph{Alguém deu um jeito no problema do visto}, meaning that someone fixed the visa problem. On the other hand, the expression \emph{dar um jeitinho}, by using the diminutive form of \emph{jeito.n}, introduces a sense of empathy and proximity. Hence, in sentences like \emph{Alguém deu um jeitinho no problema do visto}, what is being said is that someone found a non-standard, possibily illegal way of solving the visa problem. This way of solving the problem may involve a favor being granted by some authority on that matter or even the corruption of such an authority. Moreover, \newcite{Schroder_Silva_2020} demonstrate that the conceptualization of \emph{jeitinho.n} involves the idea of making rules flexible via an exchange of favors. 

Given this scenario, if we want to create a frame for \emph{jeitinho.n}, we would start by searching the Global FrameNet Database for this LU. Since there is no frame for this LU in br-pt or for any translation of it in another language, we proceed to the frame creation process. First, we select the \emph{event} root type, since, as the example sentence in the previous paragraph shows, this LU tends to occur with support verbs and indicates an action taken by someone towards solving a problem. Next, we name the frame as \texttt{Brazilian\_way} and connect it to the \texttt{Attempting\_and\_resolving\_scenario} frame in FrameNet. We will then map FE in the latter to the ones we are in the process of creating. Hence, the \textsc{Agent} FE is mapped to the \textsc{Interested\_party} and the \textsc{Goal}, \textsc{Manner} and other non-core FEs in the mother frame are repeated in the newly created frame. Because \emph{jeitinho.n} requires the conceptualization of an \textsc{Authority} being convinced – or corrupted – and of some \textsc{Norm} being violated, we add those two core FEs by clicking the \emph{Create New FE} button. Next, additional non-core FEs typycally occuring in eventive frame may be added. Finally, we edit the frame definition, and the description and coreness status of some FEs. Appendix A shows how this process is performed in Lutma step by step. 

\section{Limitations and Outlook}
\label{sec:limitations}

Despite already being deployed, Lutma is not yet a finished project. As with most software, there is room for improvement in regards to user experience and interface. Those issues will be dealt with after we receive more user feedback. There are also important functionalities that still need to be designed and developed, and because of that, in its current state, the system has some limitations.

One of them has to do with the fact that there are no tools that could aid users in assessing the quality of newly created frames. This process is not objective either, since previous research has shown how frame annotation can be ambiguous \cite{burchardt2006salsa}. Interestingly, this also means that the subsystems implemented to reduce redundancy during the frame creation process could benefit from multiple perspectives. When those subsystems fail and thus, a user creates a redundant frame, or one with overall less quality, a reasonable solution is the adoption of a wiki-like approach, where users can see edits and open discussions to determine the best course of action.

One final point worth considering is that Lutma's metalanguage is English, meaning that only users proficient in this language will be able to contribute. Naturally, this can be circumvented by allowing users to translate attributes of frames, LUs, FEs and any other entities. However, this would also mean that users would spend less time actually creating those entities. For now, we have decided to leave English as the metalanguage, taking into consideration that other factors can also restrict potential users, even though we have no control over them (\textit{e.g.} Lutma's contributors are most likely people interested in fields such as Frame Semantics or NLP).

Even given its limitations, Lutma is a step in an effort of scaling up FrameNet, without sacrificing the model's advantages, especially in regards to perspective in NLP, to the extent that it facilitates contributions from any language and from non-specialist users.

\section{Acknowledgements}

The development of Lutma was funded by an Anneliese Maier Research Award provided to Mark Turner by the Alexander von Humboldt Foundation. Torrent's research is funded by CNPq grant 315749/2021-0.

% \nocite{*}
\section{Bibliographical References}\label{reference}
%\label{main:ref}

\bibliographystyle{lrec2022-bib}
\bibliography{lrec2022-example}

% \section{Language Resource References}
% \label{lr:ref}
% \bibliographystylelanguageresource{lrec2022-bib}
% \bibliographylanguageresource{languageresource}

\onecolumn
\section*{Appendix A - User Interface Screenshots}

This section presents screenshots of Lutma's UI as presented to the user during the creation of the \texttt{Brazilian\_way} frame as described in section \ref{sec:example}. Since it is a mobile-first UI, we present how it is rendered in smartphones. The focus on mobile experience was a decision made by the team to make sure that users that do not have access to a desktop computer could also contribute. This decision also influenced the design of most screens. Some processes were split into multiple screens, which in the end, also makes it easier for unfamiliar users to work with some complex FrameNet concepts.

\begin{figure}[ht!]
    \centering
    \begin{subfigure}[t]{0.24\linewidth}
        \centering
        \includegraphics[width=\linewidth]{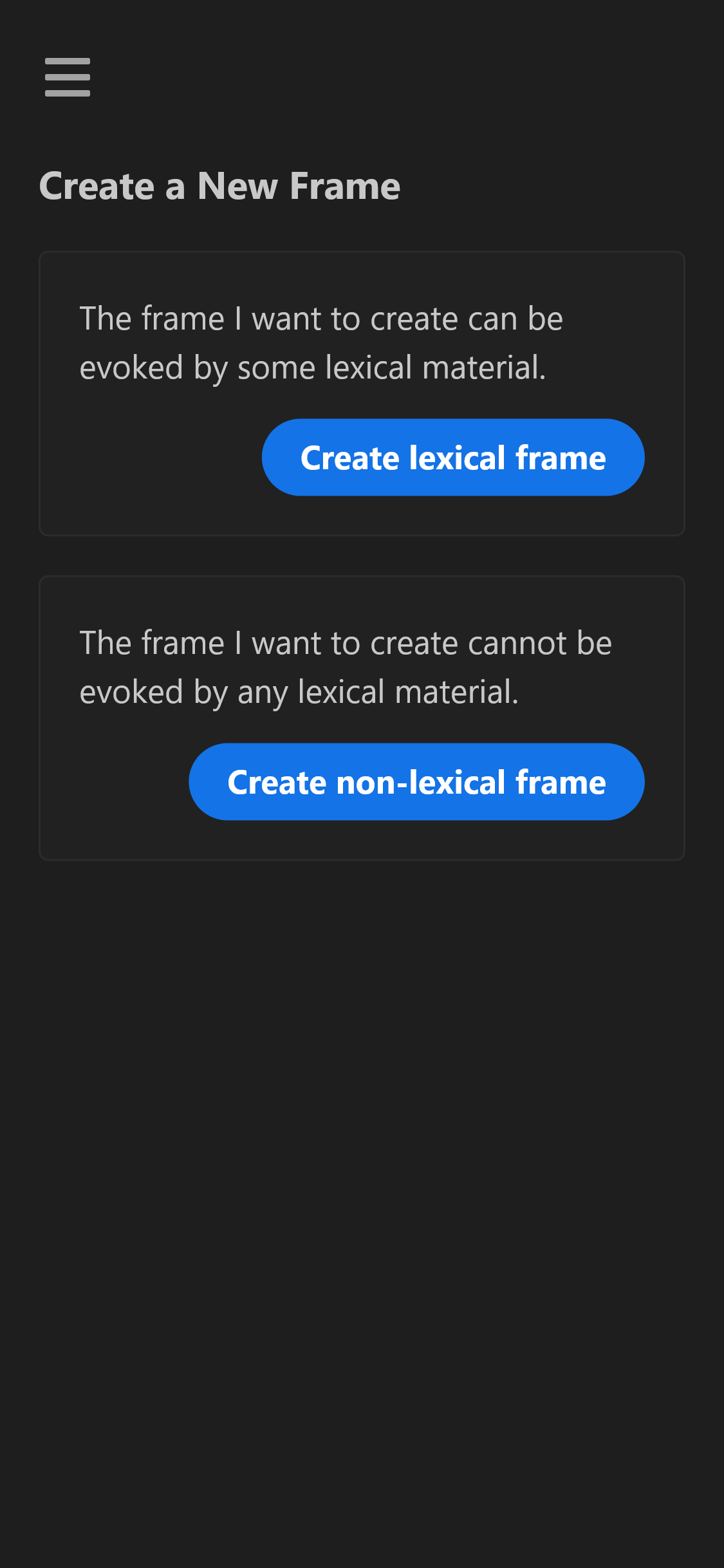}
        \caption{Frame creation screen.}
    \end{subfigure}%
    ~ 
    \begin{subfigure}[t]{0.24\linewidth}
        \centering
        \includegraphics[width=\linewidth]{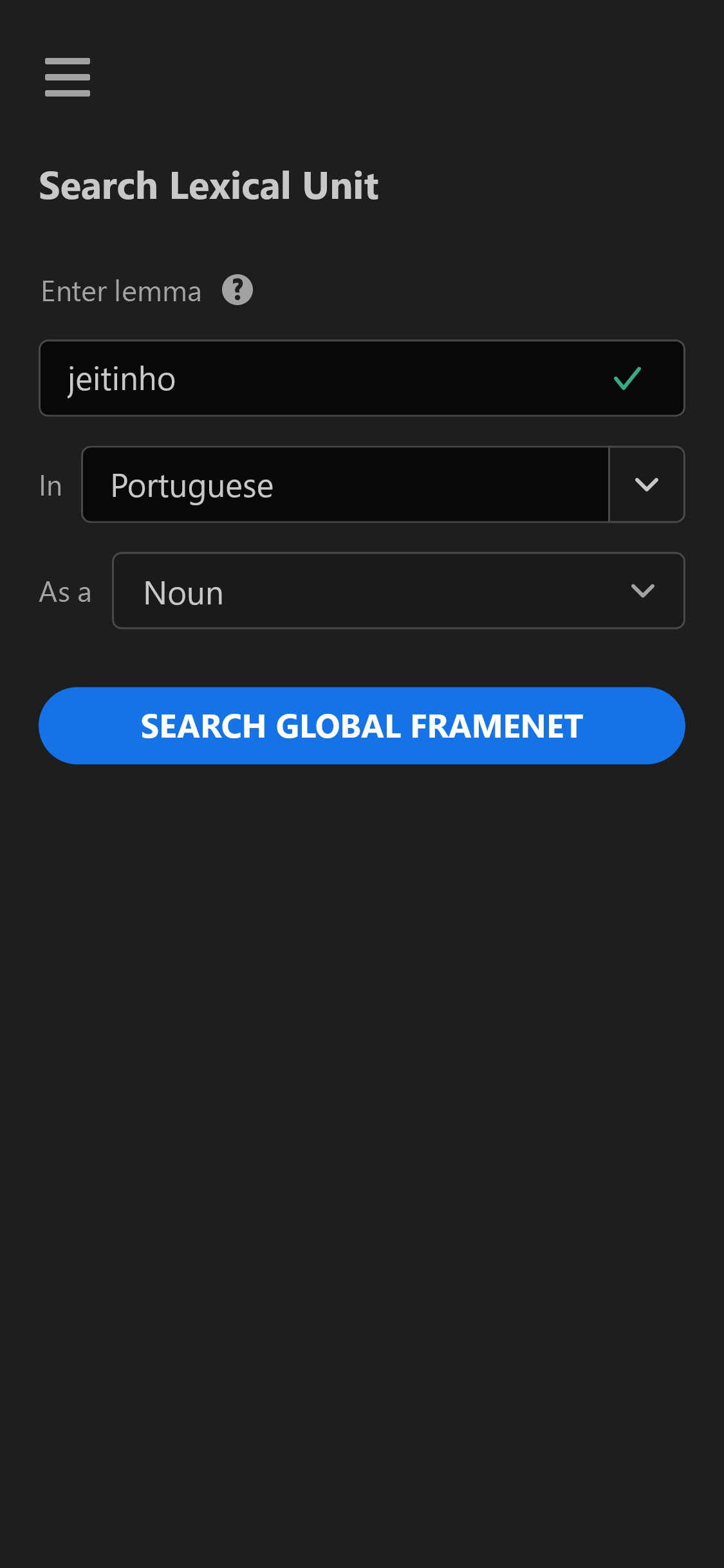}
        \caption{Lexical unit search screen.}
    \end{subfigure}%
    ~ 
    % \begin{subfigure}[t]{0.24\linewidth}
    %     \centering
    %     \includegraphics[width=\linewidth]{img/03-search-result.png}
    %     \caption{Search result screen (with a potential matching LU).}
    % \end{subfigure}%
    ~ 
    \begin{subfigure}[t]{0.24\linewidth}
        \centering
        \includegraphics[width=\linewidth]{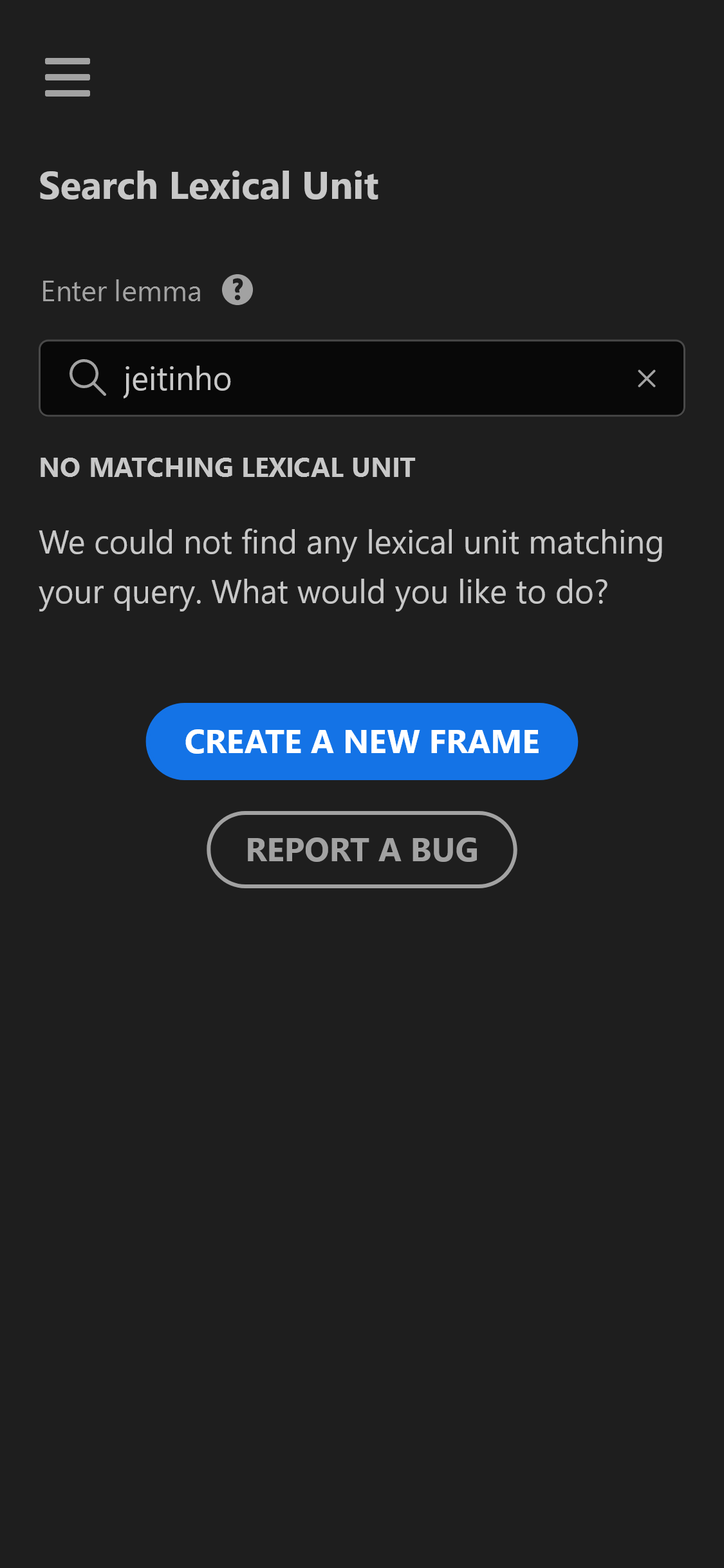}
        \caption{Search result screen after no existing frame is found.}
    \end{subfigure}%
    % \caption{Lexical unit search screens used for lexical frame creation.}
\end{figure}

\begin{figure}[ht!]
    \centering
    \begin{subfigure}[t]{0.24\linewidth}
        \centering
        \includegraphics[width=\linewidth]{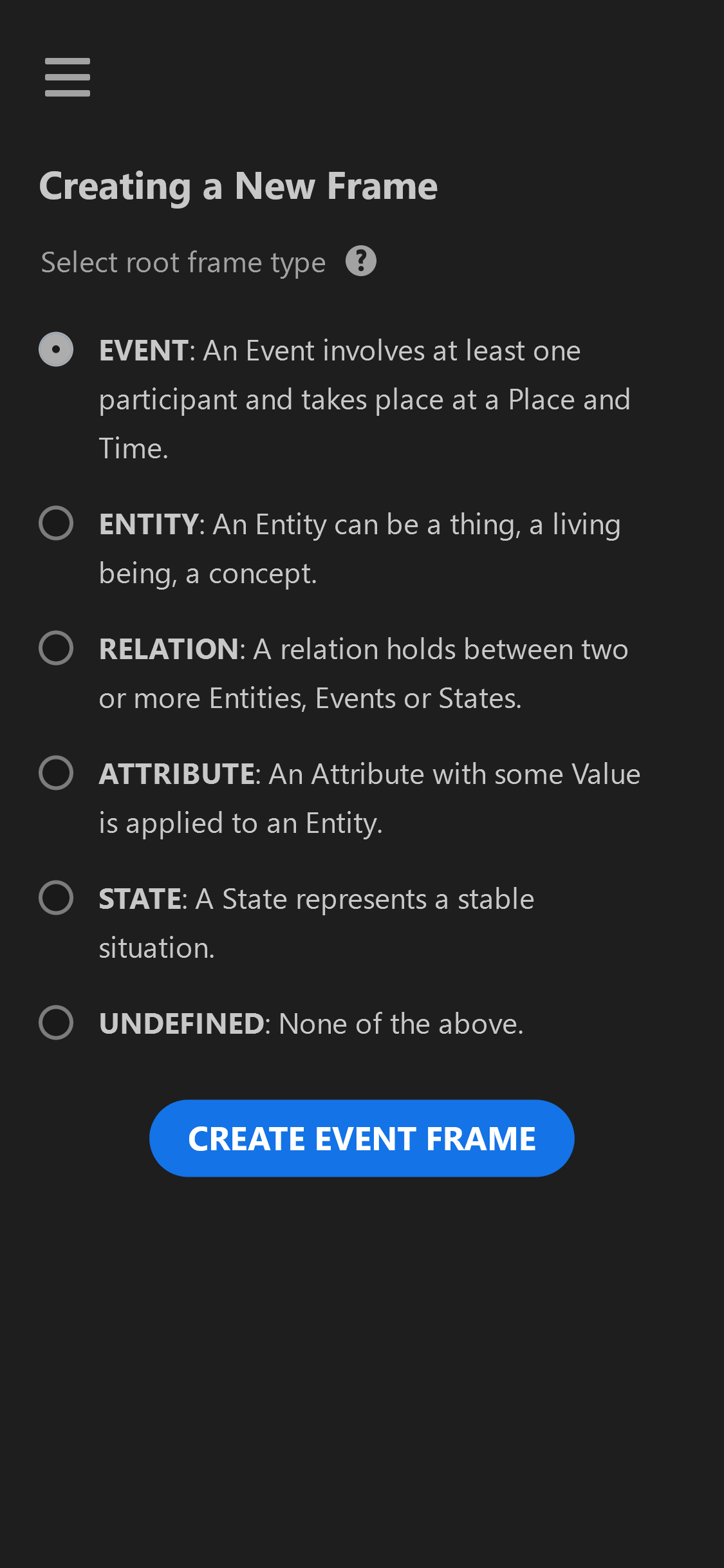}
        \caption{Frame type selection screen.}
    \end{subfigure}%
    ~ 
    \begin{subfigure}[t]{0.24\linewidth}
        \centering
        \includegraphics[width=\linewidth]{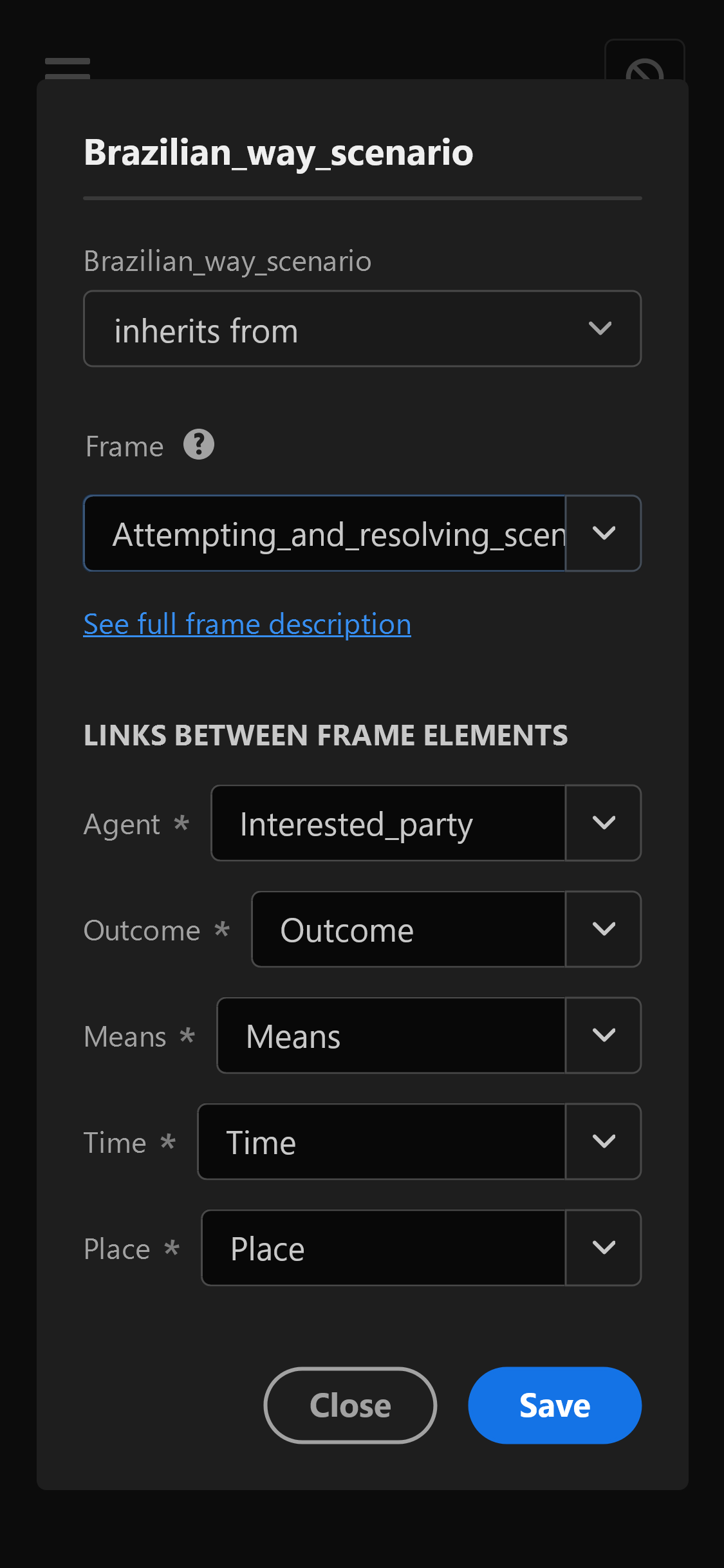}
        \caption{Frame relation creation dialog.}
    \end{subfigure}%
    % ~ 
    % \begin{subfigure}[t]{0.24\linewidth}
    %     \centering
    %     \includegraphics[width=\linewidth]{img/08-frame-relations.png}
    % \end{subfigure}%
    ~ 
    \begin{subfigure}[t]{0.24\linewidth}
        \centering
        \includegraphics[width=\linewidth]{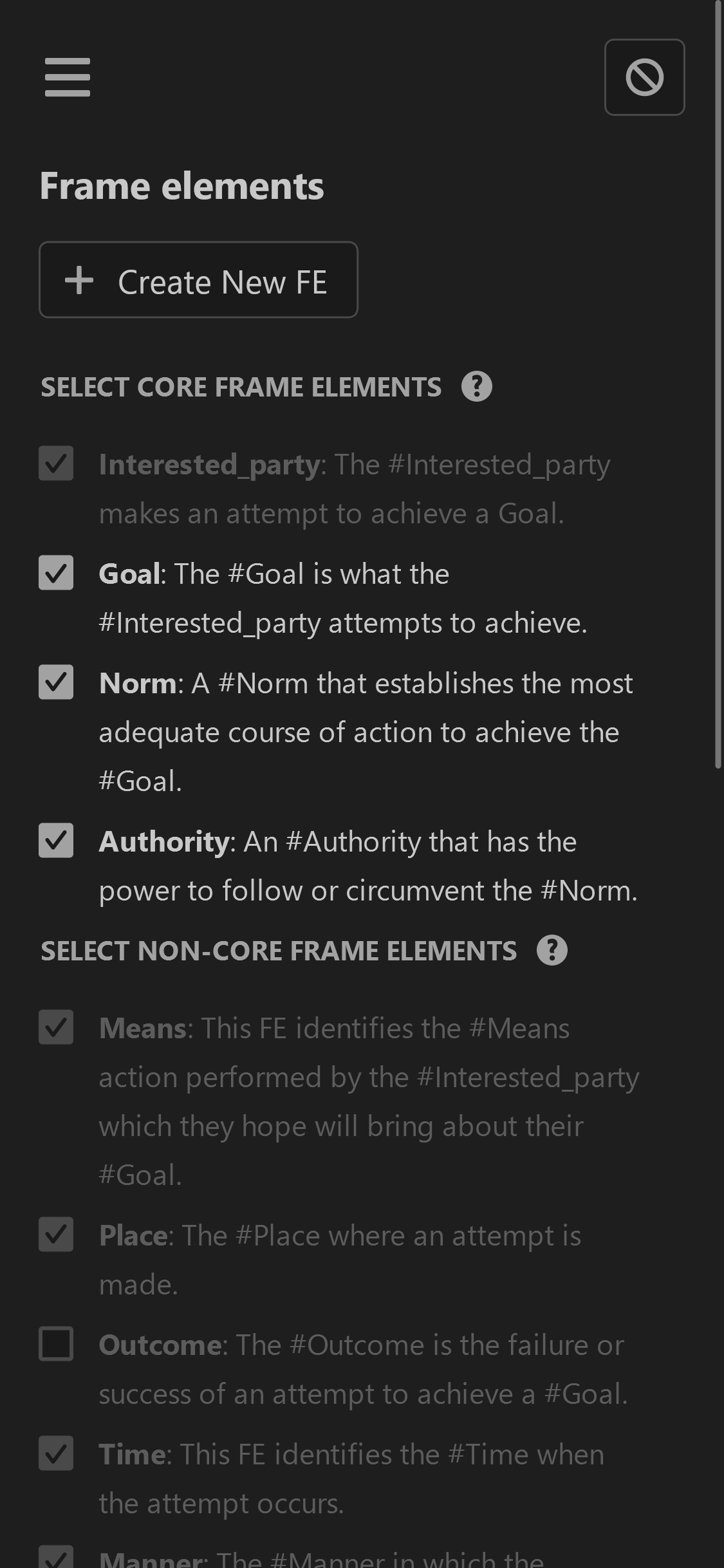}
        \caption{Frame element list screen, including non-core FEs suggested by Lutma.}
    \end{subfigure}%
    % ~ 
    % \begin{subfigure}[t]{0.24\linewidth}
    %     \centering
    %     \includegraphics[width=\linewidth]{img/10-fe-2-fe-relations.png}
    % \end{subfigure}%
    ~ 
    \begin{subfigure}[t]{0.24\linewidth}
        \centering
        \includegraphics[width=\linewidth]{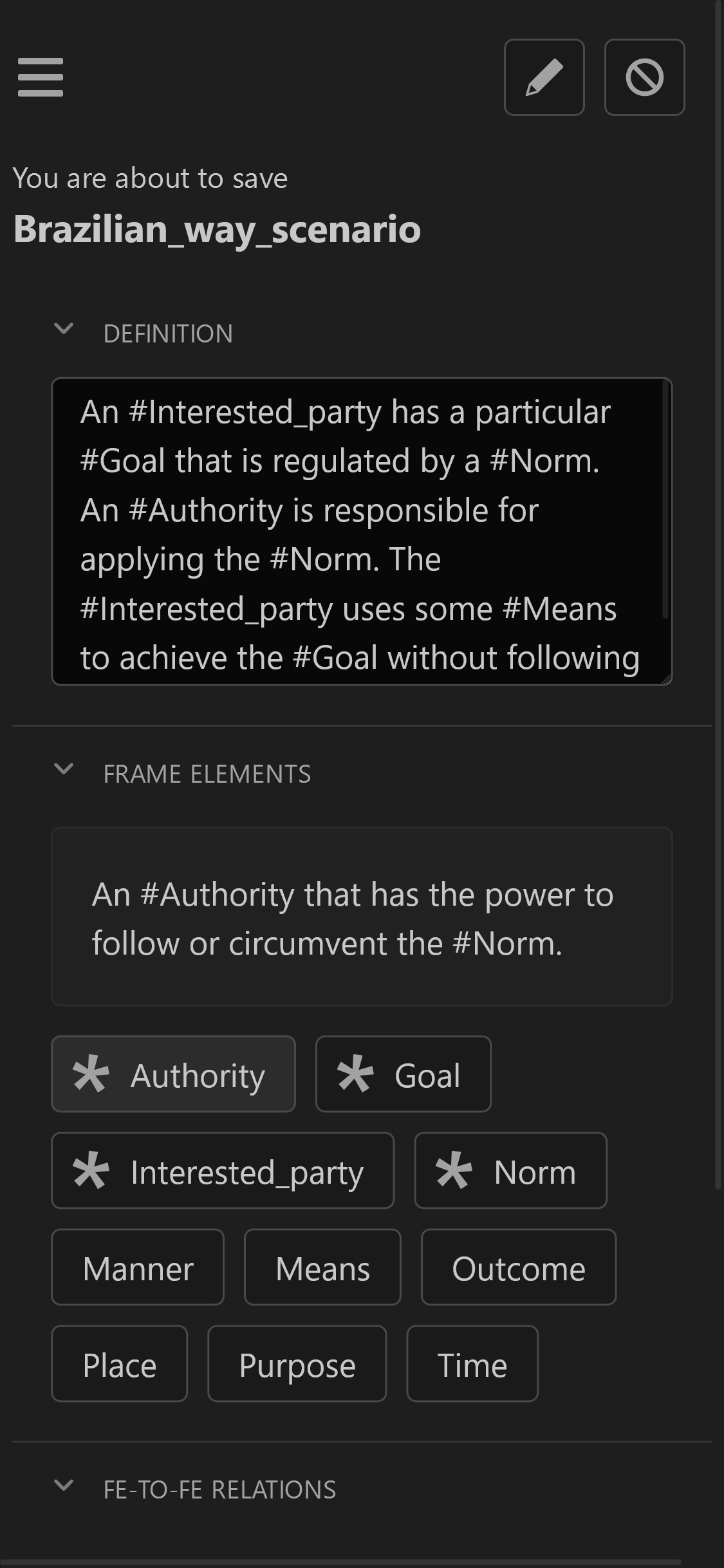}
        \caption{Summary screen displayed at the end of the execution flow.}
    \end{subfigure}%
    % \caption{Caption place holder}
\end{figure}

\end{document}